\documentclass[10pt, a4paper]{article}
\usepackage[final]{lrec_docs/lrec2026} % this is the new style
\usepackage{graphicx}
\usepackage{color}
\usepackage{xcolor}
\usepackage{booktabs}
\usepackage{float}
\usepackage{multirow}
\usepackage{subfig}
\usepackage{siunitx}
\usepackage{longtable}
\usepackage{tabularx}
\usepackage{tabularray}
\usepackage{dblfloatfix}
\usepackage{url} % or url
\usepackage[skip=3pt]{caption}
\usepackage{needspace}
\usepackage{listings}
\usepackage[table]{xcolor}

\title{Large Language Models Unpack Complex Political Opinions through Target-Stance Extraction}

\name{
\begin{tabular}{@{}c@{}}
\"Ozg\"ur Togay$^{1}$, Javier Garcia Bernardo$^{1}$, \\ 
Florian Kunneman$^{2}$, Anastasia Giachanou$^{1}$
\end{tabular}
} 

\address{$^{1}$Department of Methodology and Statistics, Utrecht University \\
        $^{2}$ Department of Languages, Literature and Communication, Utrecht University\\
         \{o.togay, j.garciabernardo, f.a.kunneman,  a.giachanou\}@uu.nl\\} 

\abstract{
Political polarization emerges from a complex interplay of beliefs about policies, figures, and issues. However, most computational analyses reduce discourse to coarse partisan labels, overlooking how these beliefs interact. This is especially evident in online political conversations, which are often nuanced and cover a wide range of subjects, making it difficult to automatically identify the target of discussion and the opinion expressed toward them. In this study, we investigate whether Large Language Models (LLMs) can address this challenge through Target-Stance Extraction (TSE), a recent natural language processing task that combines target identification and stance detection, enabling more granular analysis of political opinions. For this, we construct a dataset of 1,084 Reddit posts from \texttt{r/NeutralPolitics}, covering 138 distinct political targets and evaluate a range of proprietary and open-source LLMs using zero-shot, few-shot, and context-augmented prompting strategies. Our results show that the best models perform comparably to highly trained human annotators and remain robust on challenging posts with low inter-annotator agreement. These findings demonstrate that LLMs can extract complex political opinions with minimal supervision, offering a scalable tool for computational social science and political text analysis.
\\ \newline \Keywords{political opinions, large language models, target stance extraction}}

\begin{document}

\maketitleabstract

\section{Introduction}
Political polarization, in which individuals or groups adopt extreme political beliefs and attitudes, has become a major concern in contemporary democracies \cite{mccoy_polarization_2018,orhanRelationshipAffectivePolarization2022a,ruggeriSynthesisEvidencePolicy2024,druckmanDoesAffectivePolarization2024}. Although the mechanisms driving polarization and its spread are still debated, social media is often viewed as a key contributor \cite{suhay_polarizing_2018,bail_exposure_2018}. Platforms such as Facebook, Twitter/X, and Reddit now serve as major venues for news consumption and information exchange \cite{newman2025a}, and have been found to foster affective polarization, echo chambers, and ideological segregation \cite{barbera_tweeting_2015,bakshy_exposure_2015}. Most studies on polarization in social media have focused on binary classifications such as left and right or partisan labels such as Democrat and Republican \cite{yangQuantifyingContentPolarization2017,garimellaLongTermAnalysisPolarization2017,darwishQuantifyingPolarizationTwitter2020,brumPoliticalPolarizationTwitter2022,pengOnlineSocialBehavior2024,bojicComparingLargeLanguage2025}. A growing body of work, however, emphasizes that polarization arises from the complex interplay of opinions and beliefs \cite{dellaposta_pluralistic_2020,turner-zwinkels_affective_2023,van_noord_nature_2024}, which calls for approaches that move beyond coarse labels toward fine-grained, target-aware measurement.

Accurate identification of individual positions on specific issues, known as stance detection, is important for quantifying polarization \cite{burnham_stance_2024}. Traditional approaches assume that the target is known in advance \cite{ghosh_stance_2019} and often rely on supervised machine learning models requiring large datasets with stance labels. These models can be context-dependent and may underperform in nuanced or unseen scenarios \cite{kucukStanceDetectionSurvey2021}. 

Target-Stance Extraction (TSE) extends stance detection by jointly identifying both the target mentioned in a text and the stance expressed toward it \cite{liNewDirectionStance2023}. This removes the need for pre-defined targets and enables analysis across a broader range of topics. Early TSE architectures used two separate fine-tuned transformer models for target identification and stance classification \cite{liNewDirectionStance2023}. While this was an improvement over conventional approaches, the need for fine-tuning limited TSE’s ability to operate without pre-defined targets.

Recent advances in instruction-tuned large language models (LLMs) provide an opportunity to apply TSE with little or no task-specific training. LLMs have demonstrated strong performance in various NLP tasks, including generating synthetic data to improve stance detection \cite{wagnerPowerLLMGeneratedSynthetic2024} and matching or exceeding human annotators in sentiment and classification tasks \cite{tornbergLargeLanguageModels2024,bojicComparingLargeLanguage2025}. By unifying target identification and stance classification in a single model, LLMs can leverage large-scale pretraining to perform TSE with minimal supervision. This approach enables scalable, fine-grained analysis of political beliefs, improving our understanding of the mechanisms that drive polarization.

In this study, we evaluate the ability of LLMs to perform target-stance extraction (TSE) on complex political discussions. We focus on \texttt{r/NeutralPolitics}, a nonpartisan and strictly moderated subreddit known for longer, nuanced and evidence-based political conversations, making it a challenging setting for stance analysis. To support this, we constructed a manually annotated dataset of 1,084 posts, with a codebook that covers 138 targets alongside a semi-open ``Other \{target\}'' category. From the 1,084 annotated posts, a gold test set of 200 posts was further validated by an expert and used as the primary benchmark for LLM evaluation.\footnote{Both datasets are publicly available at \url{https://github.com/zgrtgy/llm-tse}.} Our evaluation of proprietary and open-source LLMs of varying sizes and prompting strategies shows that they can reliably identify political beliefs, with the best model achieving an F1 score of 0.76 for target identification and 0.87 for stance detection. Even when predictions differ from the gold labels, LLMs still produce reasonable, interpretable outputs. These results highlight LLMs' potential as a scalable, nuanced tool for analyzing political opinions beyond partisan proxies.

\section{Data and Methodology}
This section describes our data collection, the annotation procedure for creating a gold-standard evaluation set, and the configuration and prompting strategies used for evaluating LLMs on target-stance extraction.

\subsection{Data Collection}
Our evaluation uses posts drawn from \texttt{r/NeutralPolitics}, a nonpartisan political forum on Reddit known for evidence-based, highly moderated discussions\footnote{https://www.reddit.com/r/NeutralPolitics/wiki/index/}. Unlike the emotionally charged interactions typical of Twitter/X, this subreddit enforces strict civility and sourcing rules, making it well-suited for nuanced stance analysis.

We retrieved historical posts via the Pushshift Archives \cite{baumgartnerPushshiftRedditDataset2020} from 2005 until April 2023, when Reddit API changes temporarily halted archiving. The dataset remains publicly available and widely used in academic research \cite{mokEchoTunnelsPolarized2023,veselovskyRedditTimeCOVID2023}. Of the 578,041 comments extracted from the subreddit, 130,390 were marked as removed in the archive, meaning their text and author information were unavailable. This likely reflects subreddit moderation policies and is notably higher than the 11\% removal rate reported for Reddit overall \cite{hofmannRedditPolitosphereLargeScale2022}.

\subsection{Data Annotation}
We created an annotated evaluation dataset to cover a diverse set of political targets and issues. Three research assistants with knowledge of U.S. politics and social media were trained over three weeks. They were provided with a preliminary codebook of frequent and polarizing targets from relevant literature \cite{iyengar_origins_2019,ANES2021,Davern2024GSS}. Following the iterative open coding approach proposed by \citet{tanweerWhyDataRevolution2021}, annotators were instructed to label posts openly, allowing new targets to emerge inductively. Weekly discussions refined the codebook, resolved ambiguities, and developed a shared understanding. The final codebook includes 138 distinct targets plus an open-ended ``Other {target}'' option, significantly more granular than prior studies \cite{mohammadSemEval2016Task62016,liPStanceLargeDataset2021}. In occasional cases of multi-target posts, the target agreed upon by both annotators was selected. The full list of targets and definitions is provided as supplementary material.

To obtain posts that were both politically relevant and balanced across stance categories, our sampling strategy evolved across multiple rounds. Initial random sampling from roughly 500,000 posts returned mostly conversational, non-political comments. Continuing this approach would have made creating a politically balanced dataset extremely time-consuming. To address this, we applied a GPT-4o-mini pre-filter to identify posts with political content, roughly balanced across positive, neutral, and negative stances. Prefilter labels were not shown to annotators, ensuring unbiased labeling. Annotators received both the submission title and the full comment text and completed four rounds of training to become familiar with the task and codebook. Agreement improved as the codebook stabilized, and this refined strategy was adopted for the main annotation phase.

Three annotators labeled posts in rotating pairs, such that each post was annotated by two annotators. In total, 1,084 posts were annotated. Krippendorff’s $\alpha$ for target detection was 0.48, indicating moderate agreement  \citep{landisMeasurementObserverAgreement1977}. As $\alpha$ penalizes skewed category distributions and treats all mismatches as equally distant, it likely underestimates agreement in our setting with 138 semantically related target categories \citep{zhaoAssumptionsIntercoderReliability2013,lacyIssuesBestPractices2015}.

\begin{figure}
\includegraphics[width=\columnwidth]{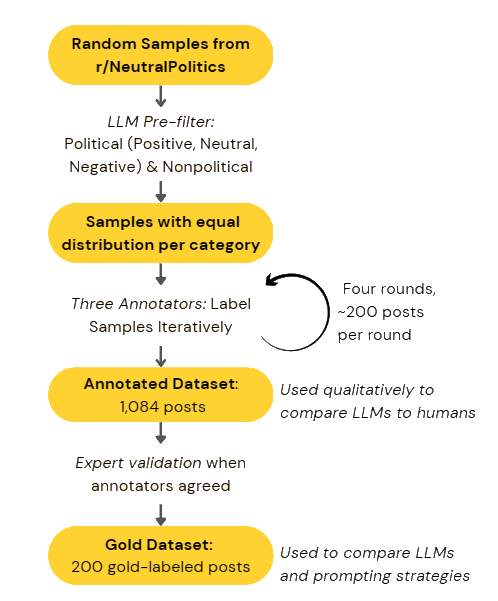}
\caption{Annotation Process}
\label{fig:annotation_flowchart}
\end{figure}

Annotators agreed on the target for 568 posts and on both target and stance for 385 posts, with disagreements most frequent in neutral stances and multi-target posts. Table~\ref{tab:agreement-distribution} presents the distribution of agreements and disagreements by stance pairs. From this agreed subset, an expert on political polarization validated a balanced sample of 200 posts (50 with no target, 50 per stance class), forming the gold-labeled dataset used for evaluation. Figure~\ref{fig:annotation_flowchart} shows the steps followed during the annotation process. This subset covers 58 distinct targets. Although smaller than the full list of 138 targets, balancing all targets across multiple stances would require oversampling rare targets, which could reduce representativeness. 
For readability, targets were grouped into higher-level categories according to their political role or type when reporting distributions. For example, ``Trump'' and ``Biden'' were assigned to \textit{Key Politicians}, ``Republicans'' and ``Democrats'' to \textit{Political Groups}, ``Christians'' and ``Hispanics'' to \textit{Ethnic \& Religious Groups}, and ``Trust in the US Military'' to \textit{Trust \& Institutions}. Table~\ref{tab:gold_target_categories} reports category-level distributions; individual target frequencies are provided in Appendix A. 
 
LLMs were evaluated against the full 138-target list, preserving the task’s granularity and difficulty. The average post length is 106 words (648 characters), considerably longer and more complex than tweet-based datasets such as PStance (30 words / 180 characters) \cite{liPStanceLargeDataset2021} and SemEval 2016 (17 words / 108 characters) \cite{mohammadSemEval2016Task62016}.

\begin{table}[!ht]
\centering
\resizebox{\columnwidth}{!}{%
\begin{tabular}{l l r r}
\toprule
\textbf{} & \textbf{Pair} & \textbf{Count} & \textbf{\%} \\
\midrule
\multirow{4}{*}{Agreements} 
 & Positive - Positive & 116 & 20\% \\
 & Neutral - Neutral   & 196 & 35\% \\
 & Negative - Negative & 73  & 13\% \\
 & \textbf{Total}      & \textbf{385} & \textbf{68\%} \\
\midrule
\multirow{4}{*}{Disagreements} 
 & Positive - Neutral  & 83  & 15\% \\
 & Positive - Negative & 34  & 6\%  \\
 & Negative - Neutral  & 66  & 12\% \\
 & \textbf{Total}      & \textbf{183} & \textbf{32\%} \\
\bottomrule
\end{tabular}%
}
\caption{Stance agreement distribution for samples with matching target labels (N = 568 of 1,084). The remaining 516 samples had mismatched targets.}
\label{tab:agreement-distribution}
\end{table}

\begin{table}[!ht]
\centering
\resizebox{\columnwidth}{!}{
\begin{tabular}{lcc}
\toprule
\textbf{Category} & \textbf{Count} & \textbf{\% of Gold Dataset} \\
\midrule
Policies \& Socioeconomic Issues & 50 & 25\% \\
Key Divisive Issues & 29 & 14.5\% \\
Key Politicians & 29 & 14.5\% \\
Trust \& Institutions & 20 & 10\% \\
Political Groups & 11 & 5.5\% \\
Technology \& Governance Issues & 5 & 2.5\% \\
International Relations & 3 & 1.5\% \\
Ethnic \& Religious Groups & 3 & 1.5\% \\
No Target & 50 & 25\% \\
\bottomrule
\end{tabular}
}
\caption{Distribution of target categories for gold-labeled posts. For mapping criteria, see Appendix A.}
\label{tab:gold_target_categories}
\end{table}

\subsection{Model Configuration}
We accessed \textit{GPT-4.1}, \textit{GPT-4.1-mini}, and \textit{o3} through OpenAI's API, while other models were accessed or deployed via Google Cloud's Vertex AI platform. Smaller open-weight models (such as \textit{Qwen} and \textit{Gemma}) were deployed directly on Vertex AI, while larger models (e.g., \textit{LLaMA}, \textit{Gemini}) were accessed via Vertex AI endpoints. Both providers explicitly state that user inputs sent through their APIs are not used for training purposes \footnote{\url{https://platform.openai.com/docs/concepts}}\footnote{\url{https://cloud.google.com/vertex-ai/generative-ai/docs/data-governance}}, mitigating concerns over data leakage.

%\footnotetext{\footnotetext{We excluded popular Deepseek open-weight models (\textit{Deepseek-V3}, \textit{Deepseek-R1}) as Deepseek does not explicitly guarantee that API-submitted data will not be retained or used for training. The smaller distilled \textit{Qwen3-8b} was locally deployable, but larger variants were impractical due to cost and infrastructure.}}

We ran all models with a fixed seed and temperature set to 0.0\footnote{Except o3, which does not support a temperature of 0.0, as it hinders reasoning.}. It should be noted that these still do not guarantee deterministic outputs; however, averages taken over three runs show minimal variation in metrics. 
%The Qwen models had reasoning disabled and used FP8 quantization, which does not affect inference quality \cite{jin-etal-2024-comprehensive,badshah2024quantifyingcapabilitiesllmsscale}.

\subsection{Prompting Strategies}
%[REFS FOR TECHNIQUES]
Prompting refers to crafting instructions that guide LLMs toward producing desired outputs. Unlike training or fine-tuning, prompting allows researchers to steer model behavior using carefully designed instructions, examples, or contextual cues.

We evaluate the most commonly used prompting strategies to assess their effect on target-stance extraction:
%[engage a lot more with the literature]

\par\textit{Zero-shot prompting:} The prompt includes only the task instruction, serving as a baseline for comparison.
\par\textit{Few-shot prompting:} Prompt includes one example per stance category to provide the model with a small reference set.
\par\textit{Conversational context augmentation:} Up to four surrounding posts from the same Reddit thread are included, prioritizing parent posts and nearby replies, to help models understand co-references and infer author intent \cite{niuChallengeDatasetEffective2024}.
\par\textit{Informational context augmentation:} Short descriptions of targets, taken directly from the annotation codebook, are appended to resolve ambiguity and standardize understanding of targets \cite{shafieiCaskowContextAwareStance2025}.

Full prompt templates are provided in Appendix C.

\section{Results}
In this section, we report the results and key findings from evaluating several LLMs under different prompting configurations on our gold-labeled dataset.

\subsection{Comparison of LLMs}
Table~\ref{tab:zero_shot} reports the performance of various LLMs on target and stance extraction using the zero-shot strategy, which serves as our baseline. Larger proprietary models (\textit{GPT-4.1, o3, Gemini-2.5}) achieve high stance detection scores (>0.80), while target identification remains around 0.70. This likely reflects the difficulty of detecting 138 possible targets, often implicitly or ambiguously mentioned in comments, compared to only three stance categories. We observe that reasoning models improve stance detection but not target identification, with sharp drops in performance for models under 8B parameters.
These results are promising given the general-purpose zero-shot setup and the challenging nature of the task.

\begin{table}[!ht]
\centering
\scriptsize
\setlength{\tabcolsep}{1pt} % tighter spacing for single-column fit
\begin{tabular}{l S[table-format=1.2] S[table-format=1.2] S[table-format=1.2] S[table-format=1.2]}
\toprule
Model & {Target F1} & {Target Acc.} & {Stance F1} & {Stance Acc.} \\
\midrule
gpt-4.1 & \textbf{0.73} & \textbf{0.70} & 0.81 & 0.81 \\
o3 & 0.71 & 0.68 & \textbf{0.84} & \textbf{0.85} \\
gemini-2.5-pro & 0.71 & 0.67 & 0.84 & 0.84 \\
gemini-2.5-flash & 0.70 & 0.66 & 0.84 & 0.84 \\
llama-3.1-405b & 0.66 & 0.66 & 0.80 & 0.80 \\
gpt-4.1-mini & 0.63 & 0.61 & 0.79 & 0.79 \\
qwen-3-32b & 0.61 & 0.58 & 0.73 & 0.73 \\
llama-3.1-70b & 0.59 & 0.60 & 0.78 & 0.78 \\
llama-4-maverick & 0.59 & 0.58 & 0.63 & 0.63 \\
qwen-3-8b & 0.58 & 0.56 & 0.72 & 0.72 \\
gemini-2.5-flash-lite & 0.57 & 0.55 & 0.74 & 0.75 \\
qwen-3-14b & 0.54 & 0.55 & 0.76 & 0.77 \\
gemma-3-27b & 0.55 & 0.50 & 0.73 & 0.73 \\
gemma-3-12b & 0.51 & 0.49 & 0.72 & 0.72 \\
llama-4-scout & 0.50 & 0.49 & 0.69 & 0.68 \\
ds-r1-qwen3-8b & 0.50 & 0.51 & 0.70 & 0.71 \\
llama-3.1-8b & 0.36 & 0.34 & 0.67 & 0.68 \\
gemma-3-4b & 0.28 & 0.27 & 0.53 & 0.57 \\
qwen-3-1.7b & 0.13 & 0.17 & 0.46 & 0.48 \\
qwen-3-0.6b & 0.04 & 0.06 & 0.35 & 0.36 \\
gemma-3-1b & 0.02 & 0.02 & 0.53 & 0.67 \\
\bottomrule
\end{tabular}
\caption{Zero-shot performance of all tested models, sorted by Target F1. Bold indicates the best score per metric.}
\label{tab:zero_shot}
\end{table}

% just describe the tables for reported models
To assess robustness to target granularity, we ran a variation in which detailed targets were replaced with broader target labels. For example, ``Republican Party,'' ``Republican politicians,'' ``Republican politician (non-listed),'' ``Republicans,'' and ``Conservatives'' were all replaced with a single label (``Republicans/Conservatives''). As shown in Table~\ref{tab:broad_label_metrics}, this slightly improves target identification, especially in larger models, while effects on stance detection are mixed. The full mapping of detailed to broader labels is provided in Appendix B.

\begin{table}[!ht]
\centering
\scriptsize
\setlength{\tabcolsep}{1pt} % tighter spacing for single-column fit
\begin{tabular}{l S[table-format=1.2] S[table-format=1.2] S[table-format=1.2] S[table-format=1.2]}
\toprule
Model & {Target F1} & {Target Acc.} & {Stance F1} & {Stance Acc.} \\
\midrule
gpt-4.1 & 0.74↑ & 0.73↑ & 0.81 & 0.82↑ \\
gemini2.5-flash & 0.74↑ & 0.70↑ & 0.86↑ & 0.86↑ \\
gemini2.5-pro & 0.72↑ & 0.68↑ & 0.85↑ & 0.85↑ \\
o3 & 0.72↑ & 0.70↑ & 0.84 & 0.85 \\
llama3-405b & 0.67↑ & 0.67↑ & 0.81↑ & 0.80 \\
gpt-4.1-mini & 0.65↑ & 0.64↑ & 0.77↓ & 0.77↓ \\
llama3.1-70b & 0.60↑ & 0.62↑ & 0.79↑ & 0.80↑ \\
llama4-maverick & 0.60↑ & 0.60↑ & 0.68↑ & 0.67↑ \\
gemini2.5-flash-lite & 0.60↑ & 0.59↑ & 0.71↓ & 0.71↓ \\
qwen3-32b & 0.57↓ & 0.56↓ & 0.76↑ & 0.77↑ \\
qwen3-14b & 0.57↑ & 0.57↑ & 0.74↓ & 0.75↓ \\
\bottomrule
\end{tabular}
\caption{TSE metrics using broad target labels, ordered by Target F1. Up arrows (↑) indicate increases from zero-shot performance, down arrows (↓) indicate decreases.}
\label{tab:broad_label_metrics}
\end{table}

\subsection{LLM Performance Across Prompting Strategies}
We next evaluate how different prompting strategies affect LLM performance, focusing on the best performing models.

\paragraph{Few-shot Prompting} Few-shot prompting consistently improves target identification across models, while stance detection remains largely stable or shows only slight gains (Table~\ref{tab:few_shot_metrics_compact}). This strategy provides strong overall performance without substantially increasing computational cost, making it an effective way to guide models with a few representative examples.

\begin{table}[!ht]
\centering
\scriptsize
\setlength{\tabcolsep}{1pt} % tighter spacing for single-column fit
\begin{tabular}{l S[table-format=1.2] S[table-format=1.2] S[table-format=1.2] S[table-format=1.2]}
\toprule
Model & {Target F1} & {Target Acc.} & {Stance F1} & {Stance Acc.} \\
\midrule
gemini-2.5-pro & 0.76↑ & 0.73↑ & 0.82↓ & 0.82↓ \\
o3 & 0.75↑ & 0.73↑ & 0.84 & 0.84↓ \\
gemini-2.5-flash & 0.74↑ & 0.71↑ & 0.81↓ & 0.82↓ \\
gpt-4.1 & 0.74↑ & 0.72↑ & 0.82↑ & 0.82↑ \\
gpt-4.1-mini & 0.70↑ & 0.68↑ & 0.78↓ & 0.78↓ \\
gemini-2.5-flash-lite & 0.61↑ & 0.59↑ & 0.78↑ & 0.78↑ \\
qwen-3-32b-fp8 & 0.63↑ & 0.61↑ & 0.72↓ & 0.72↓ \\
llama-4-maverick & 0.62↑ & 0.60↑ & 0.70↑ & 0.69↑ \\
llama-3.1-405b & 0.65↓ & 0.65↓ & 0.84↑ & 0.84↑ \\
llama-3.1-70b & 0.55↓ & 0.54↓ & 0.78↑ & 0.79↑ \\
qwen-3-14b-fp8 & 0.57↑ & 0.57↑ & 0.75↓ & 0.75↓ \\
\bottomrule
\end{tabular}
\caption{TSE metrics using few-shot prompting, ordered by Target F1. Arrows indicate changes from zero-shot performance.}
\label{tab:few_shot_metrics_compact}
\end{table}

\paragraph{Conversational Context Augmentation} Including up to four surrounding thread posts yields mixed results (Table~\ref{tab:thread_context_metrics}). Some models (e.g., GPT-4.1 variants) perform worse, suggesting sensitivity to longer prompts. Strong reasoning models (Gemini-2.5, Llama 4 Maverick, o3) show more consistent benefits, especially for target identification. These gains, however, come with substantially higher token costs.

\begin{table}[!t]
\centering
\scriptsize
\setlength{\tabcolsep}{1pt} % tighter spacing for single-column fit
\begin{tabular}{l S[table-format=1.2] S[table-format=1.2] S[table-format=1.2] S[table-format=1.2]}
\toprule
Model & {Target F1} & {Target Acc.} & {Stance F1} & {Stance Acc.} \\
\midrule
gemini-2.5-pro & 0.74↑ & 0.70↑ & 0.84 & 0.84 \\
o3 & 0.73↑ & 0.72↑ & 0.84 & 0.85 \\
gemini-2.5-flash & 0.72↑ & 0.68↑ & 0.84 & 0.84 \\
llama-4-maverick & 0.62↑ & 0.59↑ & 0.73↑ & 0.73↑ \\
gemini-2.5-flash-lite & 0.60↑ & 0.58↑ & 0.77↑ & 0.77↑ \\
llama-3.1-405b & 0.62↓ & 0.61↓ & 0.83↑ & 0.83↑ \\
llama-3.1-70b & 0.58↓ & 0.57↓ & 0.84↑ & 0.84↑ \\
qwen-3-32b-fp8 & 0.58↓ & 0.57↓ & 0.76↑ & 0.77↑ \\
gpt-4.1 & 0.66↓ & 0.64↓ & 0.78↓ & 0.79↓ \\
qwen-3-14b-fp8 & 0.50↓ & 0.52↓ & 0.83↑ & 0.84↑ \\
gpt-4.1-mini & 0.56↓ & 0.53↓ & 0.75↓ & 0.76↓ \\
\bottomrule
\end{tabular}
\caption{TSE metrics using conversational context, ordered by Target F1. Arrows indicate changes from zero-shot performance.}
\label{tab:thread_context_metrics}
\end{table}

\paragraph{Informational Context Augmentation} Providing short explanations of or background information about the targets consistently helps models identify the correct targets, as shown in Table~\ref{tab:info_context_metrics}. The effect on stance detection is less consistent.
\begin{table}[!ht]
\centering
\scriptsize
\setlength{\tabcolsep}{1pt} % tighter spacing for single-column fit
\begin{tabular}{l S[table-format=1.2] S[table-format=1.2] S[table-format=1.2] S[table-format=1.2]}
\toprule
Model & {Target F1} & {Target Acc.} & {Stance F1} & {Stance Acc.} \\
\midrule
gemini-2.5-flash & 0.75↑ & 0.71↑ & 0.82↑ & 0.83↑ \\
gemini-2.5-pro & 0.75↑ & 0.73↑ & 0.84↑ & 0.84↑ \\
gpt-4.1 & 0.74↑ & 0.73↑ & 0.81↓ & 0.81↓ \\
o3 & 0.73↑ & 0.72↑ & 0.84 & 0.85 \\
llama-3.1-405b & 0.70↑ & 0.68↑ & 0.81↑ & 0.81↑ \\
gpt-4.1-mini & 0.67↑ & 0.64↑ & 0.75↓ & 0.75↓ \\
gemini-2.5-flash-lite & 0.64↑ & 0.63↑ & 0.76↑ & 0.77↑ \\
llama-3.1-70b & 0.62↑ & 0.62↑ & 0.74↓ & 0.74↓ \\
qwen-3-32b-fp8 & 0.64↑ & 0.61↑ & 0.73 & 0.73 \\
llama-4-maverick & 0.59 & 0.58 & 0.81↑ & 0.81↑ \\
qwen-3-14b-fp8 & 0.57↑ & 0.56↑ & 0.71↓ & 0.72↓ \\
\bottomrule
\end{tabular}
\caption{TSE metrics using informational context, ordered by Target F1. Arrows indicate changes from zero-shot performance.}
\label{tab:info_context_metrics}
\end{table}

\begin{figure}
%\centering
%\includegraphics[width=1.1\columnwidth]
\includegraphics[width=1.0\columnwidth]{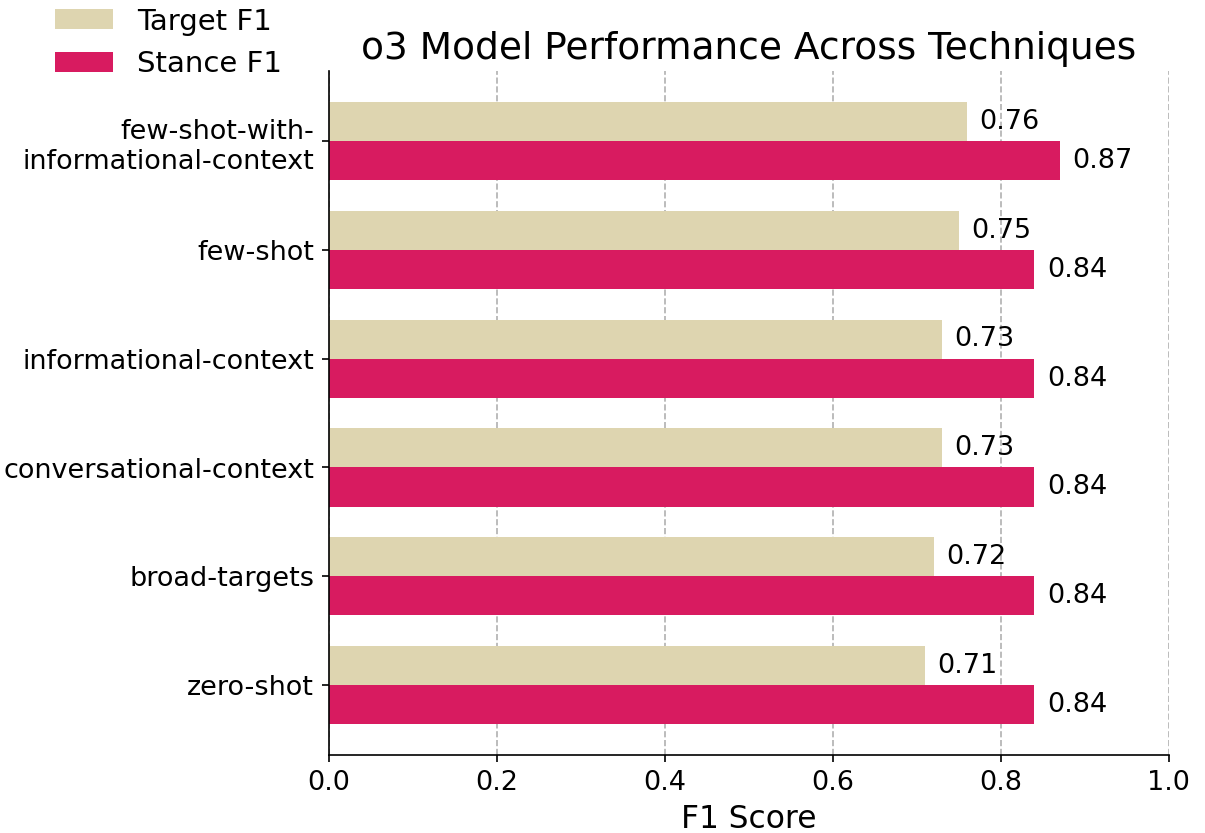}
\caption{Target and Stance F1 scores for o3 model across different prompting strategies.}
\label{fig:o3_bar_chart}
\end{figure}

\paragraph{Few-shot with Informational Context} To further enhance performance, we tested combining few-shot prompting with additional informational context (i.e., the codebook descriptions of targets) in the prompt (Table~\ref{tab:few_shots_info_context_compact}). While this does not consistently improve all models, pairing it with the o3 model yielded the best results in our tests, achieving the highest combined Target and Stance F1 scores (Figure~\ref{fig:o3_bar_chart}). 
%\textcolor{blue}{Target F1 increased by 0.05 over zero-shot and 0.01 over few-shot, while Stance F1 increased by 0.03 over both zero- and few-shot baselines.}
\begin{table}
\centering
\scriptsize
\setlength{\tabcolsep}{1pt} % tighter spacing for single-column fit
\begin{tabular}{l S[table-format=1.2] S[table-format=1.2] S[table-format=1.2] S[table-format=1.2]}
\toprule
Model & {Target F1} & {Target Acc.} & {Stance F1} & {Stance Acc.} \\
\midrule
o3 & 0.76↑ & 0.75↑ & 0.87↑ & 0.87↑ \\
gemini-2.5-pro & 0.73↓ & 0.72↓ & 0.84↑ & 0.85↑ \\
gemini-2.5-flash & 0.73↓ & 0.70↓ & 0.83↑ & 0.84↑ \\
gpt-4.1 & 0.74↑ & 0.72↑ & 0.82↑ & 0.82↑ \\
gpt-4.1-mini & 0.70 & 0.68 & 0.77↑ & 0.78 \\
gemini-2.5-flash-lite & 0.63↑ & 0.62↑ & 0.75↓ & 0.75↓ \\
llama-4-maverick & 0.63↑ & 0.61↑ & 0.68↓ & 0.67↓ \\
qwen-3-14b-fp8 & 0.58↑ & 0.57 & 0.75 & 0.75 \\
llama-3.1-70b & 0.61↑ & 0.60↑ & 0.75↓ & 0.76↓ \\
qwen-3-32b-fp8 & 0.63 & 0.61 & 0.71↓ & 0.71↓ \\
llama-3.1-405b & 0.65↓ & 0.65↓ & 0.84↓ & 0.84↓ \\
\bottomrule
\end{tabular}
\caption{TSE metrics using few-shot prompting with informational context, ordered by Target F1. Arrows indicate changes from few-shot performance.}
\label{tab:few_shots_info_context_compact}
\end{table}

Overall, our results indicate that larger, proprietary models perform best under all conditions on both target and stance detection, and that some prompting strategies, especially few-shot combined with contextual information, can noticeably improve target identification without sacrificing stance detection.

\section{Qualitative Analysis}
This section presents a qualitative analysis of the top-performing model–technique pair, o3 with few-shot prompting and informational context. We examine errors on the gold-labeled dataset and the model's behavior on posts where annotators disagreed, highlighting common misclassifications and its informative role in challenging cases.

\subsection{Error Analysis}
We manually reviewed cases where o3 predictions disagreed with gold labels. This review is interpretive and not intended as a statistical evaluation.

Two main causes of target misclassification emerged. Table~\ref{tab:examples_gold} shows three rephrased example posts where the o3 model made a prediction error. First, some posts mention multiple plausible targets, with annotators and the model sometimes emphasizing different ones. Second, o3 occasionally used titles and linked text to infer targets, while annotators focused on the post body (Ex. 2). For stance, o3 often interpreted weak cues as indicative of a stance, whereas annotators relied on stronger signals (Ex. 3).

\begin{table}[!ht]
\centering
\footnotesize
\begin{tblr}{
  colspec = {Q[c,0.05\linewidth,c] Q[c,0.40\linewidth,l] Q[c,0.35\linewidth,l] Q[c,0.15\linewidth,c]}, 
  row{1} = {font=\scriptsize\bfseries,rowsep=0pt}, 
  row{2-Z} = {font=\scriptsize}, 
  colsep = 2pt
}

% \hline
% \SetCell[c]{c} \# & Post & Labels & Note \\
% \hline

\SetCell[c=4]{c} \\
\hline
\# & Post & Labels & Note \\
\hline

1 & I’ve never had an issue with funding social programs, especially compared to the money Republicans pouring into the war machine. 
& {%
\textbf{o3:} Republican Party, Negative\\
\textbf{Gold:} Social spending, Positive%
}
& Focus on alternate target \\

\hline
2 & [Ongoing.]www.cnn.com/\linebreak
politics-flynn-russia-calls-investigation

& {%
\textbf{o3:} Belief in Trump-Russia Collusion, Neutral\\
\textbf{Gold:} None, N/A%
}
& Target inferred by URL text \\
\hline

3 & Why the Democratic Party doesn't treat campaign finance reform as a major issue?
& {%
\textbf{o3:} Democratic Party, Negative\\
\textbf{Gold:} Democratic Party, Neutral%
}
& Using weak cues \\
\hline

\end{tblr}

\caption{Example posts showing o3 prediction errors compared with gold labels. Posts are rephrased to protect anonymity.}
\label{tab:examples_gold}
\end{table}

Grouping targets into categories shows that performance is highest for more tangible or specific categories, such as \textit{Key Politicians}, and lowest for abstract or diffuse categories, such as \textit{Trust \& Institutions} (Table~\ref{tab:target_categories}). It should be noted that the sample size is too small for statistical interpretation.

\begin{table}[!ht]
\centering
\small
\resizebox{\columnwidth}{!}{
\begin{tabular}{lccc}
\toprule
\textbf{Category} & \textbf{Count} & \textbf{F1} & \textbf{Accuracy} \\
\midrule
International Relations         &  3 & 1.00 & 1.00 \\
Key Politicians                 & 29 & 0.93 & 0.90 \\
Key Divisive Issues             & 29 & 0.90 & 0.86 \\
Political Groups                & 11 & 0.88 & 0.91 \\
Policies \& Socioeconomic Issues & 50 & 0.86 & 0.82 \\
No Target                        & 50 & 0.75 & 0.60 \\
Technology \& Governance Issues &  5 & 0.67 & 0.60 \\
Trust \& Institutions           & 20 & 0.60 & 0.50 \\
Ethnic \& Religious Groups      &  3 & 0.33 & 0.33 \\
\bottomrule
\end{tabular}
}
\caption{Target identification performance across target categories.}
\label{tab:target_categories}
\end{table}

Finally, we analyzed o3's performance over stance categories when it selected the same target as annotators. We find that o3 has the most difficulty with neutral labels (Table~\ref{tab:stance_metrics}), mirroring annotators’ experience (see Table~\ref{tab:agreement-distribution}).

\begin{table}[!ht]
\centering
\footnotesize
\begin{tabular}{lccc}
\toprule
\textbf{True Stance} & \textbf{Count} & \textbf{F1} & \textbf{Accuracy} \\
\midrule
Positive  & 43 & 0.90 & 0.93 \\
Negative  & 40 & 0.84 & 0.88 \\
Neutral   & 36 & 0.76 & 0.69 \\
\bottomrule
\end{tabular}
\caption{Performance of o3 by stance (for posts with matching target).}
\label{tab:stance_metrics}
\end{table}

\subsection{Disagreement Cases}
We also evaluated o3 on posts where annotators disagreed, representing particularly challenging cases. In a random sample of 20 such posts, o3 matched the expert label in 12, produced reasonable labels in 3, and was incorrect in 5. These findings suggest that using o3 as a third annotator could often yield labels consistent with expert judgment (see Table~\ref{tab:examples_disagreement}). When errors occurred, o3 tended to overinterpret cues, inferring stronger stances (Ex. 1).

\begin{table}[!t]
\centering
\footnotesize
\begin{tblr}{
  colspec = {Q[c,0.05\linewidth,c] Q[c,0.45\linewidth,l] Q[c,0.50\linewidth,c]},
  row{1} = {font=\scriptsize\bfseries,rowsep=0pt}, 
  row{2-Z} = {font=\scriptsize}, 
  colsep = 0pt
}

%\hline
%\SetCell[c]{c} \# & \SetCell[c]{c} Post & \SetCell[c]{c} Labels \\
%\hline

\SetCell[c=4]{c} \\
\hline
\# & Post & Labels \\
\hline

1 & It’s not about raising the minimum wage. There are awful gaps where you lose your benefits if you earn more, so making less ends up better.
& {%
\textbf{o3:} Social spending, Negative\\
\textbf{A1:} Minimum wage, Negative\\
\textbf{A2:} Social spending, Neutral\\
\textbf{EX:} Social spending, Neutral%
} \\
\hline

2 & Biden sees the Green New Deal as an important foundation for tackling the climate crisis we’re facing.
& {%
\textbf{o3:} Joe Biden, Neutral \\
\textbf{A1:} Joe Biden, Neutral \\
\textbf{A2:} Green Energy, Neutral \\
\textbf{EX:} Joe Biden, Neutral%
} \\
\hline

3 & Good thing about Libertarians is they don't chase popularity. Dems and Reps need government support for influence and funding.
& {%
\textbf{o3:} Democrats, Negative \\
\textbf{A1:} Liberals, Positive \\
\textbf{A2:} Democrats, Negative \\
\textbf{EX:} Democrats, Negative%
} \\
\hline
\end{tblr}

\caption{Example posts from disagreement cases. Labels show o3 predictions, human annotator labels (A1, A2), and the expert label (EX).}
\label{tab:examples_disagreement}
\end{table}

%[discuss results]
%- mirrors other studies on LLM annotaiton
%- satisfactory results
%- in cases of disag. LLMs usually give correct answers 
% engage with other SD papers
\section{Conclusion}
This study evaluates LLM performance on Target-Stance Extraction, a key task for political opinion mining, with a focus on nuanced political discussions. It makes three primary contributions to computational social science and natural language processing.
First, we show that LLMs can unpack complex political opinions through TSE, providing an efficient way to study how different beliefs interact in polarization dynamics. Effective application, however, still requires a clear codebook, prompts and understanding of the relevant targets and literature.
Second, we present a modular and reproducible analysis framework that supports multiple prompting strategies—including zero-shot, few-shot, and context-augmented variants. This works with both proprietary and open-source models, enabling systematic comparisons across systems and configurations.
Third, we release a dataset of 1,084 Reddit comments from \texttt{r/NeutralPolitics}, 200 with gold-standard labels. It includes detailed stance and target annotations across 138 distinct political issues relevant to US politics, providing a valuable benchmark for future research on online political communication and LLM evaluation in high-context settings.
Together, these contributions demonstrate that LLMs can serve as effective tools for extracting nuanced political opinions at scale. This may help researchers to gain deeper insight into the structure and evolution of political beliefs, offering new opportunities to study polarization and discourse dynamics in online communities.

\section*{Limitations}
This study has a few limitations. First, our dataset is drawn exclusively from \texttt{r/NeutralPolitics}, a forum with complex, nuanced discussions. While this focus may limit generalizability, it also highlights the robustness of our methods in high-context, challenging texts and suggests that applying them to simpler political content could be even more straightforward. 

Second, our approach relies on a curated list of targets. Future work could explore Open-Target Stance Detection (OTSD) \citep{akashCanLargeLanguage2025}, which identifies targets in a fully open-ended manner rather than guiding the model with a predefined list. While OTSD is valuable for exploratory studies, some degree of target standardization remains necessary for practical analysis and comparability. In use cases like ours, where researchers can identify relevant targets beforehand through the literature, this is not necessary.

\section*{Data Availability}
The datasets, the codebook, the code, and other supplementary materials associated with this study are publicly available at the following link \url{https://github.com/zgrtgy/llm-tse}.

\section*{Ethical Statement}
This project received ethical approval from the Ethical Review Board of the Faculty of Social and Behavioural Sciences at Utrecht University. All example posts were rephrased to protect anonymity of users.

\section*{Acknowledgements}
This study received funding from a research grant awarded by the Applied Data Science focus area at Utrecht University. Generative AI tools were used only for formatting issues and grammatical checks.

\vspace{5mm}

\section*{References}
\label{sec:reference}

\bibliographystyle{lrec_docs/lrec2026-natbib}
\bibliography{references/references}

\appendix

\clearpage
\onecolumn
\section*{Appendix A. Gold Dataset Target Counts and Category Mappings}
\label{app:target_category_mapping}

%\begin{center}
\centering
\raggedbottom
\centering
\scriptsize
\setlength{\tabcolsep}{4pt}
\begin{tabular}[t]{p{0.25\textwidth} p{0.35\textwidth} S[table-format=2.0]}
\toprule
\textbf{Category} & \textbf{Target} & {Count} \\
\midrule
\multirow{3}{*}{\parbox{0.25\textwidth}{International Relations}} & Foreign military interventions by the US & 1 \\
 & United Nations & 1 \\
 & Israel & 1 \\
\midrule
\multirow{6}{*}{\parbox{0.25\textwidth}{Key Divisive Issues}} & Gun Control & 14 \\
 & Immigration & 5 \\
 & LGBTQ rights & 3 \\
 & Religion (general) & 3 \\
 & Belief in Climate Change & 2 \\
 & Abortion & 2 \\
\midrule
\multirow{3}{*}{\parbox{0.25\textwidth}{Ethnic \& Religious Groups}} & Muslims & 1 \\
 & Hispanics & 1 \\
 & Christians & 1 \\
\midrule
\multirow{10}{*}{\parbox{0.25\textwidth}{Key Politicians}} & Donald Trump & 12 \\
 & Hillary Clinton & 5 \\
 & Bernie Sanders & 3 \\
 & Barack Obama & 2 \\
 & G.W.\ Bush & 2 \\
 & Ron Paul & 1 \\
 & Ted Cruz & 1 \\
 & Joe Biden & 1 \\
 & Elizabeth Warren & 1 \\
 & Ronald Reagan & 1 \\
\midrule
None & None & 50 \\
\midrule
\multirow{14}{*}{\parbox{0.25\textwidth}{Policies \& Socioeconomic Issues}} & Public healthcare & 12 \\
 & Restrictions on Voting Access & 8 \\
 & Social spending & 6 \\
 & Minimum wage/Higher minimum wage & 4 \\
 & Universal basic income & 4 \\
 & International Trade & 3 \\
 & Tax cuts (general) & 3 \\
 & Student loan forgiveness & 2 \\
 & Military spending & 2 \\
 & Private healthcare & 2 \\
 & Capital Punishment & 1 \\
 & Protectionism & 1 \\
 & Police spending & 1 \\
 & Government spending & 1 \\
\midrule
\multirow{9}{*}{\parbox{0.25\textwidth}{Political Groups}} & Republican Party & 2 \\
 & Democratic Party & 2 \\
 & Black Lives Matter movement & 1 \\
 & Democratic politicians & 1 \\
 & Republicans & 1 \\
 & Republican politicians & 1 \\
 & Democratic politician (generic) & 1 \\
 & Antifa & 1 \\
 & Republican politician (generic) & 1 \\
\midrule
\multirow{4}{*}{\parbox{0.25\textwidth}{Technology \& Governance Issues}} & Net Neutrality & 2 \\
 & Use of AI in the workplace & 1 \\
 & Social Media Regulations & 1 \\
 & Increased Surveillance & 1 \\
\midrule
\multirow{8}{*}{\parbox{0.25\textwidth}{Trust \& Institutions}} & Trust in the US electoral system & 7 \\
 & Trust in the US government (general) & 4 \\
 & Trust in the US Military & 2 \\
 & Trust in the Obama administration & 2 \\
 & Trust in the US judicial system/courts & 2 \\
 & Trust in the US Supreme Court & 1 \\
 & Trust in the Trump administration & 1 \\
 & Trust in the Biden administration & 1 \\
\bottomrule
\end{tabular}
\captionof{table}{Target counts and category mappings for the gold dataset (N = 200), ordered by count descending within each category. Covers 58 of 138 codebook targets. The full target list with category mappings is provided as supplementary material.}
\label{tab:target_category_mapping}

%\end{center}

\section*{Appendix B. Detailed to Broader Labels Mappings}
\label{app:broader_label_mappings}
\begin{center}
\scriptsize
\setlength{\tabcolsep}{4pt}
\begin{tabular}{p{0.20\textwidth} p{0.20\textwidth} @{\hspace{10pt}} p{0.20\textwidth} p{0.20\textwidth}}
\toprule
\textbf{Broad Label} & \textbf{Detailed Label} & \textbf{Broad Label} & \textbf{Detailed Label} \\
\midrule
\multirow{4}{*}{\parbox{0.20\textwidth}{Republicans/Conservatives}} & Republican Party & \multirow{2}{*}{\parbox{0.20\textwidth}{Joe Biden}} & Trust in the Biden administration \\
 & Republican politicians &  & Joe Biden \\
\cmidrule(lr){3-4}
 & Republican politician (generic) & \multirow{2}{*}{\parbox{0.20\textwidth}{Donald Trump}} & Trust in the Trump administration \\
 & Republicans &  & Donald Trump \\
\midrule
\multirow{3}{*}{\parbox{0.20\textwidth}{Democrats/Liberals}} & Democratic Party & \multirow{2}{*}{\parbox{0.20\textwidth}{Barack Obama}} & Trust in the Obama administration \\
 & Democratic politicians &  & Barack Obama \\
\cmidrule(lr){3-4}
 & Democratic politician (generic) & Trust in the US Military & Trust in the US Military \\
\midrule
Antifa & Antifa & Trust in the US electoral system & Trust in the US electoral system \\
\midrule
Abortion & Abortion & Restrictions on Voting Access & Restrictions on Voting Access \\
\midrule
Gun Control & Gun Control & \multirow{2}{*}{\parbox{0.20\textwidth}{Trust in the US judicial system/courts}} & Trust in the US judicial system/courts \\
\cmidrule(lr){1-2}
LGBTQ rights & LGBTQ rights &  & Trust in the US Supreme Court \\
\midrule
Belief in Climate Change & Belief in Climate Change & Capital Punishment & Capital Punishment \\
\midrule
Immigration & Immigration & Social Media Regulations & Social Media Regulations \\
\midrule
Muslims & Muslims & Use of AI in the workplace & Use of AI in the workplace \\
\midrule
Christians & Christians & Religion (general) & Religion (general) \\
\midrule
Hispanics & Hispanics & Net Neutrality & Net Neutrality \\
\midrule
\multirow{5}{*}{\parbox{0.20\textwidth}{Progressive socioeconomic policies}} & Government spending & United Nations & United Nations \\
\cmidrule(lr){3-4}
 & Minimum wage/Higher minimum wage & Foreign military interventions by the US & Foreign military interventions by the US \\
\cmidrule(lr){3-4}
 & Social spending & Israel & Israel \\
\cmidrule(lr){3-4}
 & Public healthcare & Black Lives Matter movement & Black Lives Matter movement \\
\cmidrule(lr){3-4}
 & Student loan forgiveness & George W. Bush & George W. Bush \\
\midrule
\multirow{4}{*}{\parbox{0.20\textwidth}{Conservative socioeconomic policies}} & Tax cuts (general) & Ronald Reagan & Ronald Reagan \\
\cmidrule(lr){3-4}
 & Private healthcare & Hillary Clinton & Hillary Clinton \\
\cmidrule(lr){3-4}
 & Military spending & Bernie Sanders & Bernie Sanders \\
\cmidrule(lr){3-4}
 & Police spending & Elizabeth Warren & Elizabeth Warren \\
\midrule
Universal basic income & Universal basic income & Ted Cruz & Ted Cruz \\
\midrule
\multirow{2}{*}{\parbox{0.20\textwidth}{International Trade}} & International Trade & Ron Paul & Ron Paul \\
\cmidrule(lr){3-4}
 & Protectionism & Increased Surveillance & Increased Surveillance \\
\midrule
Trust in the US government (general) & Trust in the US government (general) & None & None \\
\bottomrule
\end{tabular}
\captionof{table}{Mapping of detailed targets to broad labels used in the broad labels prompting variant.}
\label{tab:broad_label_mapping}
\end{center}

\clearpage

\twocolumn
\raggedright
\section*{Appendix C. Prompts and Few-Shot Examples}
\label{app:prompts}
\normalsize
All prompting strategies share the same base structure. We present the
zero-shot prompt in full below, followed by a description of the
modifications applied in each variant. Table~\ref{tab:prompt-variants}
summarizes the differences.

\begin{table}[!ht]
\centering
%\scriptsize
\small
\setlength{\tabcolsep}{4pt}
\begin{tabular}{p{0.30\columnwidth}p{0.60\columnwidth}}
\toprule
\textbf{Variant} & \textbf{Change from zero-shot baseline} \\
\midrule
Few-shot & Adds \texttt{Examples} section (one per stance) before the target list \\
Broad labels & Replaces fine-grained target list with merged categories (see Appendix~B) \\
Conversational context & Restructures input format to include surrounding posts; adds \texttt{Context Weighting} section \\
Few-shot with informational context & Combines few-shot examples with per-target descriptions from the codebook \\
\bottomrule
\end{tabular}
\caption{Prompt variants and their differences from the zero-shot baseline.}
\label{tab:prompt-variants}
\end{table}

\subsection*{C.1 Zero-Shot Prompt (Baseline)}
\lstset{
  basicstyle=\small\ttfamily,
  breaklines=true,
  breakindent=0pt,
  frame=none
}
\begin{lstlisting}
You will be provided with a Reddit comment and the title of the submission
under which it was posted.

Your task is to identify the **target** and the **stance** expressed toward
it in the Comment, using the predefined list of political targets at the end.

Only classify stances related to a target from the list. If the Comment
refers to a similar target with different wording, select the exact matching
entry from the list.
**Do not rename, paraphrase, or invent new target names.**
- If the post refers to a target not on the list, label it **"Other"**.
- If no political target is mentioned, label it **"None"**.

Use the Submission **only** to resolve ambiguity (e.g., resolving pronouns
or vague references). **Do not infer stance from the Submission.**

---

## Classification Steps

1. **Identify the Political Target**
   - Choose a target from the predefined list at the end.
   - If the post refers to a target not on the list, label it **"Other"**.
   - If no target is mentioned, label it **"None"** and skip stance
     classification.

2. **Determine the Stance** (only if a target is identified)
   - **Positive**: clear support or praise
   - **Neutral**: mention without a clear stance
   - **Negative**: criticism or disapproval
   - **None**: no stance can be identified

3. **Return the classification**

---

## Output Format

Your response must be a JSON object with the following fields:
- "target" (string): The identified political target, from the list of
  Predefined Targets.
- "stance" (string): The determined stance. Must be one of: "Positive",
  "Neutral", "Negative", or "None".
- "confidence" (float): A confidence score between 0.0 and 1.0 (inclusive)
  representing how certain the model is about its classification.

Example JSON output if a target and stance are identified:
{
  "target": "Donald Trump",
  "stance": "Positive",
  "confidence": 0.95
}

---

## Predefined Targets

{predefined_targets}

(Use the exact wording. Do not rename, paraphrase, or invent targets.)
\end{lstlisting}

\subsection*{C.2 Few-Shot Variant}

The few-shot prompt is identical to the zero-shot baseline, with one
addition: an \texttt{Examples} section inserted between the output
format specification and the predefined targets list. It contains
three labeled examples, one per stance category, drawn from \texttt{r/NeutralPolitics} but held out from the annotation data. The examples are shown below.

\begin{lstlisting}[basicstyle=\small\ttfamily, breaklines=true]
## Examples

# Example 1
Input:
    Submission Title: New York Primary Results Megathread
    Comment: Didn't Sanders vote for that crime bill too? Has he evolved
    on that position or does he still support it?
Output:
{
  "target": "Bernie Sanders",
  "stance": "Neutral",
  "confidence": 0.8
}

# Example 2
Input:
    Submission Title: Flat-tax in the U.S. - a good idea?
    Comment: I'd say it's a good thing, but the reason for existing tax
    breaks is to encourage people to live in a way that is good for
    society; get educated, own property, have kids, etc.
Output:
{
  "target": "Tax cuts (general)",
  "stance": "Positive",
  "confidence": 0.90
}

# Example 3
Input:
    Submission Title: Is drug legalization/decriminalization sound policy?
    Comment: Treating drug abuse as a medical problem instead of a criminal
    problem has been successful in European countries and has been a
    conclusion by various studies.
Output:
{
  "target": "War on Drugs",
  "stance": "Negative",
  "confidence": 0.90
}
\end{lstlisting}

\subsection*{C.3 Broad Labels Variant}

The broad labels prompt is identical to the zero-shot baseline. The only
change is in the content of the \texttt{\{predefined\_targets\}} placeholder:
fine-grained targets are replaced with broader labels. The full
mapping from detailed to broad labels is provided in Appendix B.

\subsection*{C.4 Conversational Context Variant}

The conversational context prompt differs from the zero-shot baseline in two
ways. First, the input description is expanded to accommodate multiple post
types from the same thread: the comment and submission title are supplemented
with up to four surrounding posts, each tagged by type (\texttt{Submission},
\texttt{Parent}, \texttt{Focus}, \texttt{Children}, \texttt{Ancestor},
\texttt{Earlier\_Sibling}, \texttt{Later\_Sibling}). Second, a
\texttt{Context Weighting} section is added after the input description,
specifying the order in which context types should be prioritized when
resolving ambiguity:

\begin{lstlisting}[basicstyle=\small\ttfamily, breaklines=true]
## Context Weighting

When using context, prioritize in the following order:
1. Children
2. Parent
3. Ancestors
4. Earlier and Later Siblings
\end{lstlisting}

The classification instructions, output format, and target list are otherwise
unchanged. The stance is always inferred from the Focus post alone; context
is used only to resolve co-references and ambiguous phrasings.

\subsection*{C.5 Few-Shot with Informational Context}

This variant combines the additions from Section~C.2 and appends
descriptions of each target from the annotation codebook directly to the
target list. The descriptions follow the exact wording used during annotation training, and are intended to resolve ambiguity between semantically similar targets. The few-shot examples are identical to those in Section~C.2.

\end{document}